\documentclass[letterpaper]{article} 
\usepackage{aaai2026}  
\usepackage{times}  
\usepackage{helvet}  
\usepackage{courier}  
\usepackage[hyphens]{url}  
\usepackage{graphicx} 
\usepackage{natbib}  
\usepackage{caption} 
\usepackage{algorithm}
\usepackage{algorithmic}

\usepackage{booktabs}
\usepackage{amsmath}
\usepackage{multirow}
\usepackage{colortbl}
\usepackage{xcolor}
\usepackage{amssymb}
\usepackage{xspace}

\newcommand{\eg}{{\emph{e.g.}},\xspace}

\newcommand{\etal}{{\emph{et al.}}\xspace}

\usepackage{newfloat}
\usepackage{listings}
\DeclareCaptionStyle{ruled}{labelfont=normalfont,labelsep=colon,strut=off} 
\floatstyle{ruled}
\newfloat{listing}{tb}{lst}{}
\floatname{listing}{Listing}
\title{MRGeo: Robust Cross-View Geo-Localization of Corrupted Images via Spatial and Channel Feature Enhancement}

\author{
    Le Wu, Lv Bo, Songsong Ouyang, Yingying Zhu\thanks{Corresponding author.}
}
\affiliations{
    College of Computer Science and Software Engineering, Shenzhen University, China \\
    2300271073@email.szu.edu.cn, 2250271009@email.szu.edu.cn, 2400101027@mails.szu.edu.cn, zhuyy@szu.edu.cn

}

\usepackage{bibentry}
\begin{document}
\maketitle

    \begin{abstract}
    \label{abstract}
    
        Cross-view geo-localization (CVGL) aims to accurately localize street-view images through retrieval of corresponding geo-tagged satellite images. While prior works have achieved nearly perfect performance on certain standard datasets, their robustness in real-world corrupted environments remains under-explored. This oversight causes severe performance degradation or failure when images are affected by corruption such as blur or weather, significantly limiting practical deployment. To address this critical gap, we introduce MRGeo, the first systematic method designed for robust CVGL under corruption. MRGeo employs a hierarchical defense strategy that enhances the intrinsic quality of features and then enforces a robust geometric prior. Its core is the Spatial-Channel Enhancement Block, which contains: (1) a Spatial Adaptive Representation Module that models global and local features in parallel and uses a dynamic gating mechanism to arbitrate their fusion based on feature reliability; and (2) a Channel Calibration Module that performs compensatory adjustments by modeling multi-granularity channel dependencies to counteract information loss. To prevent spatial misalignment under severe corruption, a Region-level Geometric Alignment Module imposes a geometric structure on the final descriptors, ensuring coarse-grained consistency. Comprehensive experiments on both robustness benchmark and standard datasets demonstrate that MRGeo not only achieves an average R@1 improvement of 2.92\% across three comprehensive robustness benchmarks (CVUSA-C-ALL, CVACT\_val-C-ALL, and CVACT\_test-C-ALL) but also establishes superior performance in cross-area evaluation, thereby demonstrating its robustness and generalization capability.  
        
    \end{abstract}

    \begin{links}
        \link{Code}{https://github.com/WLHASH/MRGeo}
    \end{links}

    \section{Introduction}
    \label{introduction}
    
        Cross-view geo-localization (CVGL) aims to retrieve the corresponding satellite image with GPS coordinates given a street-view query image. It serves as an auxiliary means or an effective supplement for providing geo-localization information in fields such as autonomous driving \cite{autodriving1,autodriving2} and robot navigation \cite{robot_navigation} when encountering urban canyons, GPS signal degradation, or absence, showcasing significant application potential \cite{saigd,plgeo}.
    
         \begin{figure}[t]
            \centering
            \includegraphics[width=1\linewidth]{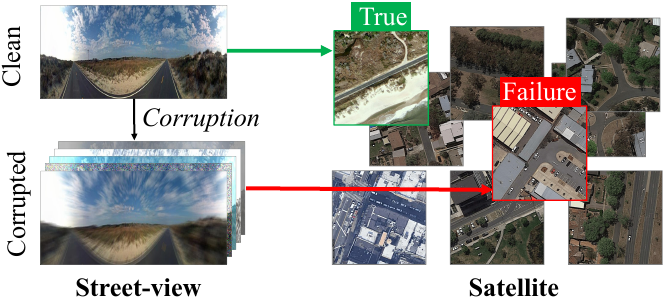}
            \caption{The fragility of existing CVGL models under real-world corruptions. While a model can easily match a clean street-view image to its correct satellite counterpart (top), its performance collapses when the query is affected by common corruptions like weather or blur, often leading to a complete failure in localization (bottom).}
            \label{fig:motivation}
        \end{figure}
    
        In recent years, with the advancement of CVGL, existing methods \cite{EPBEV} achieved near-perfect performance on some standard datasets (\eg CVUSA). However, an undeniable fact is that these datasets are collected under clean conditions and do not fully reflect the common corrupted scenes (\eg rain) prevalent in the real world. This distribution shift between training data and real-world data causes a drastic performance degradation or even failure when models trained on clean images are applied to corrupted data, limiting the reliability and robustness of models in practical applications \cite{benchmarke_zhang}, as illustrated in Fig.~\ref{fig:motivation}. Therefore, enhancing model robustness in corrupted environments is crucial for building truly reliable and deployable CVGL systems. 
        
        To comprehensively and fairly assess the robustness of CVGL methods under corruption, Zhang \etal first proposed a robustness benchmark for CVGL \cite{benchmarke_zhang}. However, to the best of our knowledge, no prior work has proposed a systematic method to address the performance degradation of CVGL methods on corrupted images. To narrow down this gap, we investigate the impact of corrupted images on feature representation capabilities and conclude that the impact of corruption on feature representation is multi-level. Specifically, in the spatial domain, fine-grained CVGL on clean images heavily relies on fine-grained local details (\eg textures). Yet, these details are the most fragile and unreliable features under corruption \cite{robust_1,robust_fourier}. Conversely, global semantic information (\eg road layouts) is more robust but often too coarse to distinguish between visually similar but geographically distinct locations. In the channel domain, corruption introduces spatially heterogeneous perturbations at the pixel level. These perturbations accumulate through the network \cite{channel_1}, distorting channel-wise information \cite{channel_2,channel_3}. But standard feature extractors and attention mechanisms, often employing a one-size-fits-all approach to channel re-weighting, are ill-suited to handle such complex, multi-granularity distortions \cite{senet,channel-wise}.

        To address the aforementioned challenges, this paper introduces MRGeo, a systematic, hierarchical method designed to counteract the significant performance degradation of CVGL models under image corruption. Our key insight is that achieving true robustness necessitates a two-level strategy: 1) enhancing the intrinsic quality of core features and 2) imposing a strong geometric structural prior.
    
        Specifically, in the spatial domain of features, the corruption creates a critical conflict: corruptions like blur primarily destroy fine-grained local details, while leaving coarse global structures relatively intact. Conversely, corruptions like fog and snow can obscure larger regions, degrading global semantic information. In the channel domain of features, corruptions such as contrast or JPEG directly perturb the pixel values, leading to distorted channel-wise statistics and a subsequent loss of semantic information. 
        
        To counteract this degradation at the feature level, we propose the Spatial-Channel Enhancement Block (SCEB). This block comprises two synergistic sub-modules engineered to fundamentally improve core feature quality:
        1) Spatial Adaptive Representation Module (SARM) as a dynamic arbitration mechanism that explicitly models global semantics and local details in parallel. It features a learned gating system that acts as a reliability estimator for local features, adaptively suppressing their contribution in the presence of corruption while leveraging them in clean conditions. 2) We introduce a Channel Calibration Module (CCM) to counteract channel-level information loss. Moving beyond simplistic channel-wise scaling, CCM performs a more sophisticated compensatory calibration. It achieves this by decoupling and modeling multi-granularity channel dependencies, allowing it to dynamically correct for channel distortions at spatial location with a richer, more comprehensive context. However, feature-level enhancement alone is insufficient. Under severe corruption, features can become highly ambiguous, causing the model to erroneously match semantically similar but geographically disparate regions. To mitigate the risk of spatial misalignment, we introduce the Region-level Geometric Alignment Module (RGAM). Based on the inherent spatial correspondence between the two views, RGAM partitions the feature map into a fixed grid and concatenates the resulting regional features in a consistent, predefined order. This structural constraint compels the model to perform matching within corresponding coarse-grained geographic areas, which fundamentally ensures geometric consistency, thereby reinforcing the robustness of the entire localization system. Our main contributions are as follows:
        \begin{itemize}
    
            \item To the best of our knowledge, this paper proposes a novel and systematic method for the performance degradation of CVGL methods on corrupted images, significantly enhancing the model's robustness and generalization under corruption. 
    
            \item To address the multi-level impact of corruption on feature representation, a hierarchical defense strategy is proposed. At the feature level, \textbf{SARM} enhances spatial representations by dynamically arbitrating between global and local information, while \textbf{CCM} calibrates channel-wise distortions to counteract information loss. At the structural level, \textbf{RGAM} imposes a rigid geometric prior, ensuring robust matching even under severe corruption.
    
            \item Extensive experiments on CVGL robustness benchmark and standard datasets comprehensively demonstrate that MRGeo not only achieves excellent performance on robustness benchmark but also achieves superior performance on cross-area tasks, fully demonstrating its effectiveness and robustness.
        
        \end{itemize}

    \section{Related Work}
    \label{related_work}
    
        \subsection{Cross-View Geo-Localization}
            Early CVGL research on datasets like CVUSA \cite{CVUSA} and CVACT \cite{CVACT} focused on improving retrieval accuracy using Siamese networks \cite{SAFA}, attention mechanisms \cite{l2ltr}, and Transformers \cite{transgeo}. Subsequent efforts addressed geometric complexities in datasets such as VIGOR \cite{vigor} with sophisticated techniques like BEV rectification \cite{EPBEV} and advanced sampling \cite{sample4geo}. While these methods progressively pushed performance boundaries, they shared a common trait: a heavy reliance on precise, fine-grained visual cues. This pursuit of precision has inadvertently created a critical vulnerability, as these fragile cues are easily compromised by common image corruptions.
        
        \subsection{Corrupted Image Robustness in CVGL}
            The performance degradation of models on corrupted images was a known challenge in computer vision \cite{robust_1}. In the context of CVGL, this issue was first systematically quantified by Zhang \etal \cite{benchmarke_zhang}, who constructed robustness benchmarks and revealed drastic performance drops across existing methods. However, their work diagnosed the problem without offering a methodological solution. To fill this critical gap, we introduce MRGeo, a novel and systematic method designed to enhance the intrinsic robustness of CVGL models. Unlike prior work focused on geometric precision, MRGeo directly confronts feature degradation caused by corruption, making it a crucial step towards reliable real-world deployment.
    
    \begin{figure*}[t]
        \centering
        \includegraphics[width=0.985\linewidth]{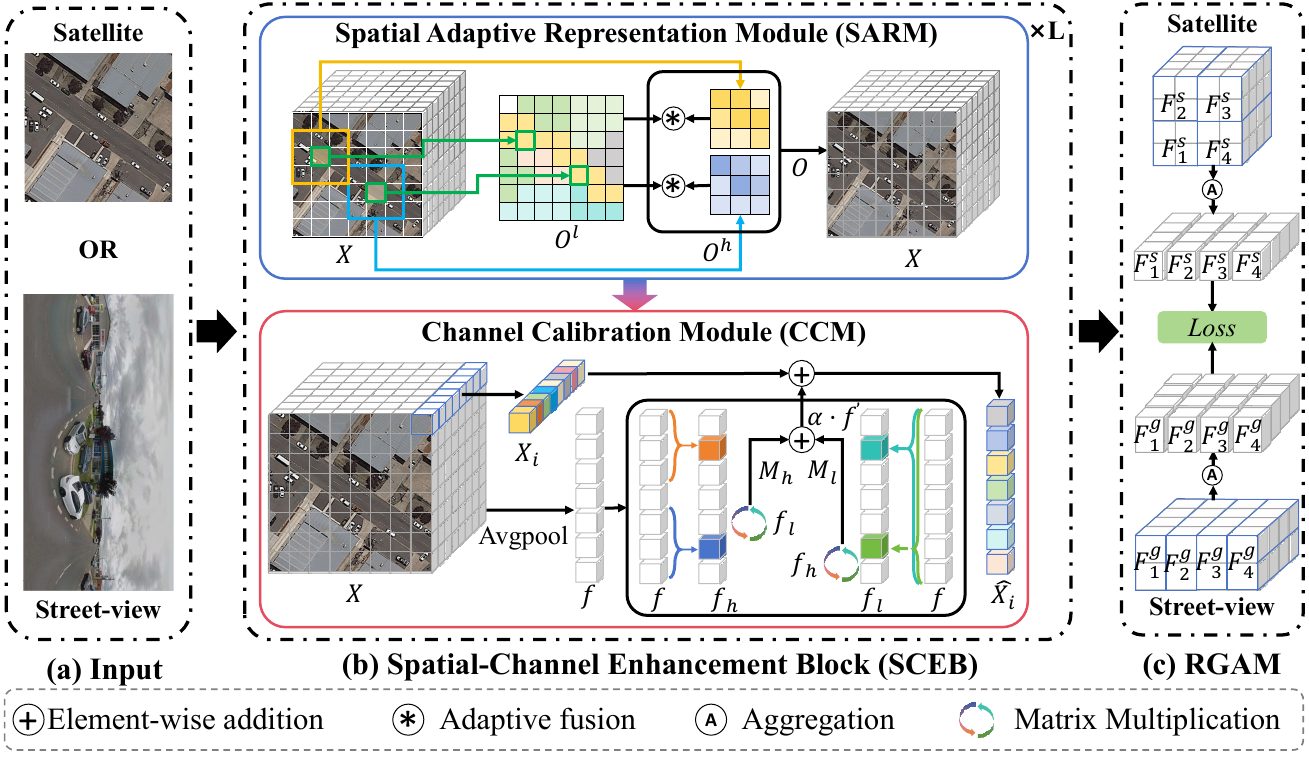}
        \caption{Overview of our MRGeo architecture. The framework processes street-view and satellite images through a shared-weight backbone containing our proposed \textbf{Spatial-Channel Enhancement Block (SCEB)}. SCEB enhances feature quality via its two sub-modules: the \textbf{SARM} and the \textbf{CCM}. Finally, the \textbf{Region-level Geometric Alignment Module (RGAM)} imposes a structural constraint on the enhanced features to generate robust final descriptors for retrieval.}
    
        \label{fig:model_overview}
    \end{figure*}

    \section{Method}
    \label{method}
    
        \subsection{Problem Definition}
        \label{problem_definition}
            Given a set of street-satellite image pairs $\{I^g_i , I^s_i \} ^N_{i=1}$, where $N$ denotes the number of image pairs, $g$ and $s$ represent the street-view and satellite respectively. Assume the image encoders $f^s$ and $f^g$ are trained on the ideal distribution $\mathcal{D}$. In the ideal scenario, we find matching pairs by minimizing the distance between features from the two views, expressed as:
            
            \begin{equation}
                \begin{aligned}
                    \mathcal{D} \sim \{ d(f^g(I^g_i), f^s(I^s_i)) < d(f^g(I^g_i), f^s(I^s_j)) \} , i \ne j
                \end{aligned}
            \end{equation}
            
            where $d(\cdot)$ denotes the $\mathcal{L}_2$ distance. However, real-world data is often affected by various corruptions $\mathcal{C}$ (\eg rain, snow, blur, noise), causing the data distribution to deviate from $\mathcal{D}$ and follow a more complex corrupted data distribution $\mathcal{P}$. On corrupted data, models trained on distribution $\mathcal{D}$ experience a significant performance degradation or even failure. Therefore, this paper primarily studies the task of accurate localization on corrupted data, and formulated it as:
            
            \begin{equation}
                \begin{aligned}
                    \mathcal{P} \sim \{ d(f^g(\mathcal{C}(I^g_i)), f^s(I^s_i)) < d(f^g(\mathcal{C}(I^g_i)), f^s(I^s_j)) \}
                \end{aligned}
            \end{equation}
    
        \subsection{Spatial Adaptive Representation Module}
        \label{samrm}
    
            The Spatial Adaptive Representation Module (SARM) is designed to resolve a fundamental conflict in robust feature extraction: corruption primarily degrades fine-grained local details, while robust global semantics remain relatively stable. To address this, SARM treats them as distinct, complementary information streams and by introducing a mechanism for dynamic, content-aware arbitration.
            
            Specifically, for an input feature map $X \in \mathbb{R}^{C \times H \times W}$, where $C$,$ H$, and $W$ denote the number of channels, height, and width respectively, we first capture global semantic structure $O^l \in \mathbb{R}^{C \times H \times W}$, which are less affected by noise or local occlusions. This is achieved via a standard self-attention mechanism where each spatial feature vector $x_i \in \mathbb{R}^C$ attends to all other positions in the map. The process is formulated as:
            
                \begin{equation}
                    \begin{aligned}
                         O^l_i &= \mathcal{S}(x_i, X) \cdot (X W_v^l)\\
                        \mathcal{S}(x_i, X) &= (x_i W_q^l) \cdot (X W_k^l)^T
                    \end{aligned}
                \end{equation}
    
            \noindent Here, $W_q^l$, $W_k^l$, and $W_v^l$ are learnable projection matrices. $\mathcal{S}(x_i, X)$ computes the similarity scores. To extract corruption-sensitive, fine-grained local context $O^h \in \mathbb{R}^{C \times H \times W}$, we apply a similar attention mechanism but restrict the interaction of $x_i$ to its local $k \times k$ neighborhood, denoted as $\mathcal{N}_k(x_i) \in \mathbb{R}^{k^2 \times C}$:
            
                \begin{equation}
                    \begin{aligned}
                        O^h_i = \mathcal{S}(x_i, \mathcal{N}_k(x_i)) \cdot (\mathcal{N}_k(x_i) W_v^h) \\
                      \mathcal{S}(x_i, \mathcal{N}_k(x_i)) = (x_i W_q^h) \cdot (\mathcal{N}_k(x_i) W_k^h)^T
                    \end{aligned}
                \end{equation}
                
            where $W_q^h, W_k^h, W_v^h$ are the corresponding learnable matrices. Finally, the $O^l$ and $O^h$ are aggregated using a gating mechanism that acts as a learned reliability estimator to form the output feature $O$:
            
                \begin{equation}
                    O = O^l \circledast O^h = O^l + \sigma(FC([O^l, O^h])) \odot O^h
                \end{equation}
            
            Here, $\circledast$ denotes an adaptive fusion operation, $[O^l, O^h]$ denotes channel-wise concatenation of the $O^l$ and $O^h$. A gating vector is generated by passing the concatenated features through a fully connected layer ($FC$) and a sigmoid activation function ($\sigma$). This gate, with values in $[0,1]$, dynamically scales the contribution of the local features $O^h$ via element-wise multiplication ($\odot$) before they are added to the global features $O^l$. This adaptive fusion allows the model to learn a dynamic trade-off: in the presence of corruption, it can suppress the unreliable $O^h$ by producing low gate values and rely on the robust $O^l$. Conversely, for clean images, it can leverage $O^h$ to achieve higher precision. 
            
            This gating mechanism is conceptually analogous to those in LSTMs and GRUs, where the model learns to control the flow of information based on the input context. Here, the ‘‘context" is the level of corruption implicitly encoded in the concatenated features $[O^l, O^h]$. The sigmoid function ensures the gate values are in a stable $[0,1]$ range, acting as a soft switch that dynamically arbitrates between global robustness and local precision. The effectiveness of this design is validated in our ablation studies (Tab \ref{table:ablation}, \#5-\#7).

        \subsection{Channel Calibration Module}  
        \label{ccm}
        
            While SARM primarily focuses on spatial robustness, the channel calibration module (CCM) is designed to mitigate the multi-faceted impact of corruption on feature channels. Previous studies have shown that not all channel information is highly relevant to the task \cite{partchannel, partchannel_1, bghr}, but high-level channel semantic features can enhance the model's discriminative ability \cite{cricavpr,cbam}. Therefore, we aim to enhance key channel representations by constructing fine-grained global channel dependencies and dynamically adjusting channel representations at spatial location, thereby strengthening the expression of key channels.
    
            The process begins by extracting a global channel representation $f \in \mathbb{R}^C$ from the input feature $X$ via global average pooling. To explore the dynamic, complementary relationships between channels, we decouple $f$ into two components. Specifically, we use a 1D convolutional layer (to capture local dependencies between adjacent channels) and a linear layer (to capture the overall pattern of global channels) to process $f$, obtaining global structural features $f_l$ and local detail features $f_h$, respectively. Then, we calculate their correlation to achieve mutual strengthening and correction. The specific process is as follows:
            
            \begin{equation}
                \begin{aligned}
                    f' = \underbrace{\mathcal{B}(f_h \cdot f_l^T)}_{\mathcal{M}_h} + \underbrace{\mathcal{B}(f_l \cdot f_h^T)}_{\mathcal{M}_l}
                \end{aligned}
            \end{equation}
            
            Here, the matrix multiplication (\eg $f_l \cdot f_h^T$) computes a cross-channel correlation matrix, where each element quantifies the interaction strength between global structural channel and local detail channel. The function $\mathcal{B}$ performs a summation along the horizontal axis of this matrix, transforming the correlation scores into a vector. The intermediate terms, $\mathcal{M}_h \in \mathbb{R}^C$ and $\mathcal{M}_l \in \mathbb{R}^C$, represent the mutual enhancement between the structural ($f_l$) and detail ($f_h$) channel features. In essence, this step allows the global channel patterns to calibrate the local channel responses, and vice-versa, leading to a more refined and robust channel representation. 
            
            The final result, a highly informative and robust channel correction signal $f'$, is used to perform a dynamic residual correction on the feature map at spatial location. After an initial projection of the input features $X$ to $\hat{X} \in \mathbb{R}^{C \times H \times W}$, we calibrate each feature vector $\hat{x_i}$ as follows: $\hat{x_i}' = \hat{x_i} + \alpha f'$, where $\alpha$ is a learnable parameter controlling the injection strength of $f'$. By making $\alpha$ learnable, the network can autonomously determine the optimal degree of calibration needed at different layers and for different tasks, avoiding a manually tuned hyperparameter. It allows global channel statistics to enhance salient features and suppress noise at each spatial location, effectively mitigating the channel-level perturbations caused by corruption and improving overall feature robustness.

        \subsection{Region-level Geometric Alignment Module}
        \label{rgam}
        
            Inspired by prior work, we recognize the importance of geometric consistency between views for the CVGL task. To further address the feature loss and interference caused by corruption, we introduce the region-level geometric alignment module (RGAM). As illustrated in Fig. \ref{fig:model_overview}(c), the RGAM operates on the feature map $X$ from SCEB. It first spatially partitions $X$ into a 2 $ \times$ 2 grid for satellite (and a 1 $\times $ 4 grid for street-view) into four non-overlapping regions ($F^i_1, F^i_2, F^i_3, F^i_4$), where the superscript $i \in \{g,s\}$ indicates the view type (street-view or satellite). Then each region is aggregated through average pooling $\mathcal{A(\cdot )}$. Finally, these regional vectors are concatenated in a fixed order to construct the final descriptor $f^i$, formulated as:
            
            \begin{equation}
                f^i = [ \mathcal{A} (F^i_1),\mathcal{A}(F^i_2), \mathcal{A}(F^i_3), \mathcal{A} (F^i_4) ]
            \end{equation}
            
            This strategy enforces a consistent spatial layout, ensuring that matching occurs between corresponding local regions. By providing robust structural cues, this approach effectively counters the loss of fine-grained details caused by corruption and enhances overall model robustness.

        \begin{table*}[h]
            \centering
            \small
            \setlength{\tabcolsep}{1.5mm}{
                \begin{tabular}{c|c|cccc|cccc|cccc}
                \toprule
                \multirow{2}{*}[-0.5ex]{\textbf{Method}} &\multirow{2}{*}[-0.5ex]{\textbf{Publication}} & \multicolumn{4}{c|}{\textbf{CVUSA-C-ALL}} & \multicolumn{4}{c|}{\textbf{CVACT\_val-C-ALL}} & \multicolumn{4}{c}{\textbf{CVACT\_test-C-ALL}} \\
                \cmidrule(lr){3-14}
                 && R@1 & R@5 & R@10 & R@1\% & R@1 & R@5 & R@10 & R@1\% & R@1 & R@5 & R@10 & R@1\% \\
                \midrule
                L2LTR       & NIPS'21 & 87.93 & 95.45 & 97.01 & 99.01 & 82.13 & 93.34 & 94.93 & \underline{98.10} & 57.20 & 82.59 & 87.23 & \underline{98.09} \\
                TransGeo    & CVPR'22 & 82.72 & 91.95 & 94.03 & 97.92 & 74.04 & 86.19 & 89.10 & 94.98 & 52.18 & 74.35 & 78.99 & 95.03 \\
                GeoDTR      & AAAI'23 & 84.64 & 93.29 & 95.01 & 98.24 & 77.40 & 88.95 & 91.28 & 95.91 & 52.87 & 78.84 & 83.17 & 95.84 \\
                Sample4G    & ICCV'23 & \underline{93.36} & 97.32 & 98.08 & 99.22 & 86.71 & 94.35	&95.60	&98.04	&66.33	& 88.05	& 90.75	&98.07\\
                EP-BEV  & ECCV'24 & 84.33 & 94.38 & 96.06 & 98.39 & 82.07 & 92.19 & 94.09 & 97.52 & 59.25 & 83.12 & 87.02 & 97.49 \\
                DReSS      & JPRS'25 & 92.42 & \underline{97.50} & \underline{98.33} & \underline{99.25} & \underline{87.39} & \underline{94.44} & \underline{95.61} & 98.00 & \underline{68.12} & \underline{88.81} & \underline{91.24} & 98.07 \\
                \textbf{MRGeo} & --  &\textbf{95.09}	 &\textbf{97.99}	 &\textbf{98.48}	 &\textbf{99.38}	 &\textbf{90.45}	 &\textbf{96.50}	 &\textbf{97.33}	 &\textbf{98.91}	 &\textbf{71.16}	 &\textbf{92.24}	 &\textbf{94.21}	 &\textbf{98.80}\\
                \bottomrule
                \end{tabular}}
            \caption{Experimental results of cross-view geo-localization methods on comprehensive corruption robustness benchmarks.}
          \label{table:cor_all}
        \end{table*}

        \begin{table*}[t] 
            \centering
            \small
            \setlength{\tabcolsep}{1.5mm}{
            \begin{tabular}{c|c|ccccc|cccc|ccc|c}
                \toprule
                \multirow{3}*[-1.5ex]{\textbf{Method}} &\multirow{3}*[-1.5ex]{\textbf{Clean}}& \multicolumn{13}{c}{\textbf{CVACT\_val-C}}\\
                \cmidrule(lr){3-15}
                &&\multicolumn{5}{c|}{\textbf{Weather}} &\multicolumn{4}{c|}{\textbf{Blur}} & \multicolumn{3}{c|}{\textbf{Digital}}\\
                \cmidrule(lr){3-7} \cmidrule(lr){8-11} \cmidrule(lr){12-14} \cmidrule(lr){15-15}
                && Snow & Frost & Fog & Bright & Spatter & Defocus & Glass & Motion & Zoom & Contrast & Pixel & JPEG & R@$1_{cor}$ \\ 
                \midrule

                L2LTR & 84.89 & 71.03 & 77.93 & 83.50 & 81.17 & 73.78 & 83.98 & 85.07 & 84.00 & 49.79 & \textbf{79.15} & 85.07 & 83.40 & 78.16 \\
                TransGeo & 84.95 & 47.65 & 58.51 & 32.91 & 72.67 & 67.13 & 81.43 & 84.83 & 81.80 & 36.34 & 22.18 & 84.92 & 83.74 & 62.84 \\
                GeoDTR & 86.21 & 48.24 & 71.74 & 83.26 & 84.60 & 61.39 & 79.11 & 85.51 & 73.44 & 8.26 & 55.48 & 86.01 & 85.19 & 68.52 \\
                Sample4G & 90.81 &78.69 &84.68 &86.97 &88.80 &84.97 &89.17 &90.22 &89.03 &\underline{50.00} &51.99 &90.26 &89.52 &81.20  \\
                EP-BEV & 88.91 & 69.43 & 78.67 & 81.55 & 86.08 & 79.77 & 87.52 & 88.55 & 87.36 & 38.15 & 49.09 & 88.77 & 86.92 & 76.82 \\
                DReSS & \underline{91.32} & \underline{81.23} & \underline{86.50} & \underline{88.14} & \underline{89.97} & \underline{85.48} & \underline{90.02} & \underline{90.89} & \underline{90.12} & 49.98 & 52.78 & \underline{90.95} & \underline{89.95} & \underline{82.17} \\
                \textbf{MRGeo} &\textbf{92.67}	&\textbf{88.75} 	&\textbf{90.97} 	&\textbf{91.25} 	&\textbf{91.95} 	&\textbf{90.22} 	&\textbf{91.47} 	&\textbf{92.35} 	&\textbf{91.79} 	&\textbf{65.31} 	&\underline{72.71} 	&\textbf{92.34} 	&\textbf{91.67} 	& \textbf{87.56} \\
    
                \bottomrule
            \end{tabular}}
            \caption{ Experimental results of 7 cross-view geo-localization methods on the CVACT\_val-C datasets. We report the R@1 performance of each method under different corruptions (obtained by averaging the 5 corruption severities), as well as the average R@$1_{cor}$ under all corruption types.}
            \label{table:CVACT_VAL-C}
        \end{table*}
        
        \subsection{Optimization Objective}
        \label{loss}
        
            Prior studies \cite{sample4geo} have established the effectiveness of contrastive learning for cross-view geo-localization. Based on this, we employ the InfoNCE loss~\cite{infonce1,infonce} with refined formulation:
            
            \begin{equation}
                \mathcal{L}_{\text{InfoNCE}}(q, R) = -\log\left[ \frac{\exp\left( \mathbf{q}^\top \mathbf{r}_+ / \tau \right)}{\sum_{i=0}^{K} \exp\left( \mathbf{q}^\top \mathbf{r}_i / \tau \right)} \right]
                \label{eq:nce_loss}
            \end{equation}
            
            where $\mathbf{q} \in \mathbb{R}^D$ denotes the $\ell_2$-normalized query embedding from street-view, $R = \{\mathbf{r}_+, \mathbf{r}_1, ..., \mathbf{r}_K\}$ represents a batch containing one positive satellite embedding $\mathbf{r}_+$ and $K$ negative samples $\mathbf{r}_i$, with $\tau > 0$ being a temperature hyperparameter. The loss minimizes the negative log-likelihood of identifying the correct geospatial correspondence, effectively pulling $\mathbf{q}$ toward $\mathbf{r}_+$ in the embedding space while pushing it away from all $\mathbf{r}_i$.

    \section{Experiment}
    \label{experiment}
    
        \subsection{Datasets and Evaluation Metrics}
        \label{dataset}
            \textbf{Datasets.} We evaluate on CVGL databases—CVUSA \cite{CVUSA}, CVACT \cite{CVACT}—and their robust variants: the fine-grained robustness benchmarks and the comprehensive robustness benchmark \cite{benchmarke_zhang}. All datasets contain precisely center-aligned street-view-satellite image pairs. The fine-grained robustness benchmarks (CVUSA-C and CVACT-C) include 16 different corruption types, each with 5 severity levels (totaling 80 evaluation sets), to systematically evaluate model robustness; the comprehensive robustness benchmark (CVUSA-C-ALL, CVACT\_val-C-ALL, and CVACT\_test-C-ALL) aggregate all corruption types into a single evaluation set. \textbf{Metrics.} Following previous work \cite{saigd}, retrieval performance is evaluated by R@K (K $\in$ \{1, 5, 10, 1\% \}), which represents the probability of correctly identifying the matching image within the top K retrieved reference images based on the query image. To specifically quantify robustness, we also report the average R@1 across all 16 corruption types, denoted as R@$1_{cor}$.

        \subsection{Implementation Details}
            
            MRGeo is designed based on ViT architecture and pretrained on LVD-142M, with an output feature dimension of 768. After processing by RGAM, the final output dimension is 3072 (4$\times$768). For optimization, consistent with prior work \cite{EPBEV}, we use the AdamW optimizer \cite{adam} combined with a cosine decay learning rate scheduler (initial learning rate $1e^{-4}$). The hyperparameter $k$ for the SARM is set to 3, a value justified by our ablation studies (see Section 4.4). Training is conducted on 4 NVIDIA V100 GPUs with a batch size of 16 for 20 epochs, with the first epoch serving as a warm-up phase.

        \subsection{Comparison with Existing Methods}
      
            We benchmark our method against 6 state-of-the-art (SOTA) methods, including L2LTR \cite{l2ltr}, TransGeo \cite{transgeo}, GeoDTR \cite{geodtr}, Sample4Geo \cite{sample4geo}, EP-BEV \cite{EPBEV}, and DReSS \cite{dress}. The comparison is performed on CVUSA-C, CVACT\_val-C, CVUSA-C-ALL, CVACT\_val-C-ALL, and CVACT\_test-C-ALL datasets. To ensure a fair and direct comparison, results for EP-BEV and DReSS are based on official public code on these benchmarks. 
    
            \begin{table}[t]
                \centering
                \small
                \setlength{\tabcolsep}{1.2mm}{
                \begin{tabular}{c|ccc|cccc}
                    \toprule
                    \multirow{2}{*}[-0.5ex]{\textbf{\#}} &\multirow{2}{*}[-0.5ex]{\textbf{SARM}}
                     &\multirow{2}{*}[-0.5ex]{\textbf{CCM}}
                     &\multirow{2}{*}[-0.5ex]{\textbf{RGAM}}
                     & \multicolumn{4}{c}{\textbf{CVACT\_test-C-ALL}}\\
                    \cmidrule(lr){5-8} 
                    &&&& R@1 & R@5  & R@10 & R@1\%  \\
                    \midrule
                    1&$\times$        &$\times$       &$\times$          	&65.84 	&88.32 	&91.03 &98.30 \\
                    2&$\times$        &$\times$       &$\checkmark$      	&67.86 	&89.90 	&92.26 &98.45 \\
                    3&$\checkmark$    &$\checkmark$	&$\times$          	&68.90 	&90.76 	&92.99 &98.56 \\
                    4&$\checkmark$    &$\checkmark$	&$\checkmark$       &\textbf{71.16} 	&\textbf{92.24}  &\textbf{ 94.21} &\textbf{98.80} \\
                    \midrule
    
                    5&2:1             &$\checkmark$   &$\checkmark$    	&70.40 	&91.37 	&93.38 &98.50\\
                    6&1:1             &$\checkmark$   &$\checkmark$    	&69.16 	&90.82 	&93.86 &98.58 \\
                    7&1:2             &$\checkmark$   &$\checkmark$    	&59.55 	&83.51 	&86.91 &97.24 \\
                  \midrule
                    8&$\checkmark$    &FC	    &$\checkmark$       &53.59 	&79.59 &84.12 &97.18 \\ 
                    9&$\checkmark$    &XCiT	&$\checkmark$       &69.87 	&90.89 &93.09 &98.62 \\
                    10&$\checkmark$    &MFCM	&$\checkmark$       &59.56 	&82.54 &85.72 &96.32 \\
            
                    \bottomrule
                \end{tabular}}
                \caption{The Ablation study of Components.}
                \label{table:ablation}
            \end{table}
            
            \paragraph{Comprehensive Corruption Robustness Evaluation.}We evaluate MRGeo's overall robustness against corruption on the comprehensive benchmarks, with the results presented in Tab \ref{table:cor_all}. MRGeo establishes a new state-of-the-art (SOTA), achieving the best performance across all three datasets. Specifically, it surpasses the SOTA method (DReSS) with R@1 performance gains of 2.67\% on CVUSA-C-ALL and 3.06\% on CVACT\_val-C-ALL. This superiority is particularly evident on the more challenging CVACT\_test-C-ALL dataset, where our method achieves a 3.04\% improvement. These results demonstrate that MRGeo can effectively handle diverse types of corruption in the real world, significantly outperforming prior methods.
            
            \paragraph{Fine-grained Corruption Robustness Evaluation.} To conduct a detailed analysis of model's performance under different corruption types, we focus our discussion on the CVACT\_val-C benchmark, whose complex scenes provide a more compelling testbed for robustness. The experimental results, presented in Tab \ref{table:CVACT_VAL-C}, show that ‘‘Snow'', ‘‘Contrast'', and ‘‘Zoom'' corruptions have the most significant impact on model performance. Even under these most impactful corruptions, our method demonstrates remarkable robustness, surpassing the SOTA method (DReSS) with performance gains of 7.52\% on ‘‘snow'' and 15.33\% on ‘‘Zoom''. 
            
            Notably, our performance on ‘‘Contrast'' warrants a closer look. While achieving a substantial 19.93\% gain compared to DReSS, its average performance is sub-optimal. This is because at severity level 5, extreme contrast reduction creates low-variance features, which flattens SARM's attention distributions and renders its adaptive gate ineffective. Concurrently, the resulting homogenization of channel-wise activations provides no distinct patterns for CCM to model, thus failing to generate a meaningful calibration signal to enhance key features. Nevertheless, our method still secures competitive performance in this type of corruption.

            \begin{table}[t]
                \centering
                \small
                \setlength{\tabcolsep}{1.5mm}{
                    \begin{tabular}{c|ccc|ccc}
                    \toprule
                    \multirow{2}*[-0.5ex]{\textbf{Method}} & \multicolumn{3}{c|}{\textbf{CVUSA $\rightarrow$ CVACT}}   & \multicolumn{3}{c}{\textbf{CVACT $\rightarrow$ CVUSA}} \\ 
                    \cmidrule(lr){2-4} \cmidrule(lr){5-7}
                    & R@1   & R@5     & R@1\%  & R@1   & R@5  & R@1\%\\
                    \midrule
                    SAFA$^\dagger$ & 30.40 & 52.93 & 85.82  & 21.45 & 36.55 & 69.83 \\
                    DSM$^\dagger$  & 33.66 & 52.17 & 79.67 & 18.47 & 34.46 & 69.01 \\
                    TransGeo  & 37.81 & 61.57 & 89.14 & 18.99 & 38.24 & 88.94 \\
                    L2LTR$^\dagger$  & 52.58 & 75.81 & 93.51 & 37.69 & 57.78 & 89.63 \\
                    GeoDTR$^\dagger$  & 53.16 & 75.62 & 93.80 & 44.07 & 64.66 & 90.09 \\
                    Sample4G  & 56.62 & 77.79 & 94.69 & 44.95 &64.36 & 90.65 \\
                    OR-CVFI$^\dagger$  & 68.07 & 83.25 & -- & 40.13 & 56.24 & -- \\
                    \midrule
                    \textbf{MRGeo} &&&&&&\\
                    $1 \times 1$ & 82.05 	&93.09 	&97.67 	&43.65 	&62.29 	&88.23 \\
                    $3 \times 3$ &81.57 	&92.84 	&97.56 	&\textbf{47.73} 	&\textbf{65.65} 	&\textbf{90.85} \\
                    $5 \times 5$ &\textbf{82.73} 	&\textbf{93.21} 	&\textbf{97.92} 	&45.00 	&63.38 	&88.63 \\
                   \bottomrule
                \end{tabular}}
                \caption{Cross-area evaluation when trained on the CVUSA dataset and evaluated on CVACT and vice versa. $\dagger$ denotes models that use the polar transformation. -- indicates that the corresponding data are not provided. The bottom section shows the impact of the hyperparameter $k$ in SARM on cross-area evaluation.}
                \label{table:cross-area_generalization}
            \end{table}
    
        \subsection{Ablation Studies}
    
            To ascertain the effectiveness of MRGeo's components, we conduct the following ablation experiments: 1) the hyperparameter $k$ of SARM; 2) the adaptive fusion operation $\circledast$ of SARM; 3) the effectiveness of CCM; and 4) the effectiveness of RGAM.
    
            \paragraph{Hyperparameter $k$ of SARM.} We find that the selection of the hyperparameter $k$ in SARM is highly correlated with scene complexity. Consequently, we use the cross-area evaluation to best illustrate this dependency. As shown in Tab \ref{table:cross-area_generalization}, when $k=1$, the extracted local detail features are too fine-grained, causing local details to suppress the representational capacity of global structural features. When $k=3$, it balances the capture of local details and global context perception well on the CVACT dataset, avoiding over-reliance on local details or omission of global information, thereby achieving optimal performance on the CVACT$\rightarrow$CVUSA task. When $k=5$, it achieves optimal performance on the CVUSA$\rightarrow$CVACT task because the CVUSA dataset has fewer scene categories and sparser unique local detail features; a larger receptive field can more effectively capture and understand local feature distributions.
            
            \paragraph{Adaptive fusion operation $\circledast$ of SARM.} To verify the effectiveness of adaptive fusion operation $\circledast$, we set the ratio of $O^l$ and $O^h$ to a fixed value. As shown in Tab \ref{table:ablation} (\#4, \#5, \#6, and \#7), as the proportion of $O^h$ increases, the proportion of global semantics $O^l$, which is more robust to corruption, decreases, and model's performance also declines. In contrast, with adaptive fusion, the ratio changes with the increase of network depth, which aligns with the findings of existing studies \cite{channel_diff_1}, indicating MRGEO has different feature preferences in regions of different depths.

            \paragraph{Effectiveness of the CCM.} We replace our channel calibration module with FC \cite{fc}, XCiT \cite{xcit}, and MFMC \cite{cricavpr} respectively. The experimental results, shown in Tab \ref{table:ablation} (\#4, \#8, \#9, and \#10), indicate that MFMC, as a multi-scale method, performs better than simple FC, demonstrating the importance of high-level channel features for enhancing feature representation capabilities. XCiT considers the positional information of different tokens and interacts with the channel information of different tokens, also achieving some improvement. Our method, by dynamically adjusting channel weights based on global channel features that are more robust to corruption, more effectively compensates for channel information perturbations and losses caused by corruption, significantly enhancing feature robustness.
            
            \paragraph{Effectiveness of the RGAM.} As shown in Tab \ref{table:ablation} ( \#1, \#2, \#3, and \#4), we verify the effectiveness of the region-level geometric alignment module. The introduction of this module improved R@1 by 2.02\% and 2.26\%, respectively. This indicates that enforcing consistency of regional features between views can effectively enhance the model's robustness when facing corruption, further narrowing the performance gap with the ideal scenario. Although its reliance on center-alignment is a limitation, an ablation study on the VIGOR dataset confirms its importance, showing a performance drop from 77.99\% to 73.03\% upon its removal.

            Furthermore, comparing experiments in Tab \ref{table:ablation} (\#2, \#3, and \#4) highlights a crucial synergistic effect. While adding only RGAM (\#2) or only the SARM+CCM block (\#3) improves performance, the complete model (\#4) achieves a disproportionately larger gain (R@1 increases by 5.32\% over the baseline). This super-additive boost confirms that our modules are complementary: the SARM and CCM tackle corruption at the feature level (spatial and channel), while the RGAM enforces geometric consistency at the descriptor level. This demonstrates that our holistic method, addressing the problem from multi-level, is substantially more effective than addressing any single aspect in isolation.

             \paragraph{Cross-area Evaluation.} To validate the model's generalization capability, we conduct a cross-area evaluation, with the results summarized in Tab \ref{table:cross-area_generalization}. In the CVUSA $\rightarrow$ CVACT task, our method achieves a remarkable 13.5\% R@1 improvement over the previous SOTA. Conversely, in the challenging CVACT$\rightarrow$CVUSA task, our model still surpasses the SOTA by 2.78\%. The performance difference stems from the datasets' characteristics. Training on the simpler CVUSA encourages the model to learn highly generalizable geometric features, leading to a substantial performance leap on the complex CVACT. Conversely, training on CVACT causes the model to learn intricate, scene-specific cues that are less transferable to CVUSA's simpler context. Crucially, MRGeo's robustness ensures it still outperforms existing methods even when learning from complex data, underscoring its superior generalization capability.

            \begin{figure}[t]
                \centering
                \includegraphics[width=0.95\linewidth]{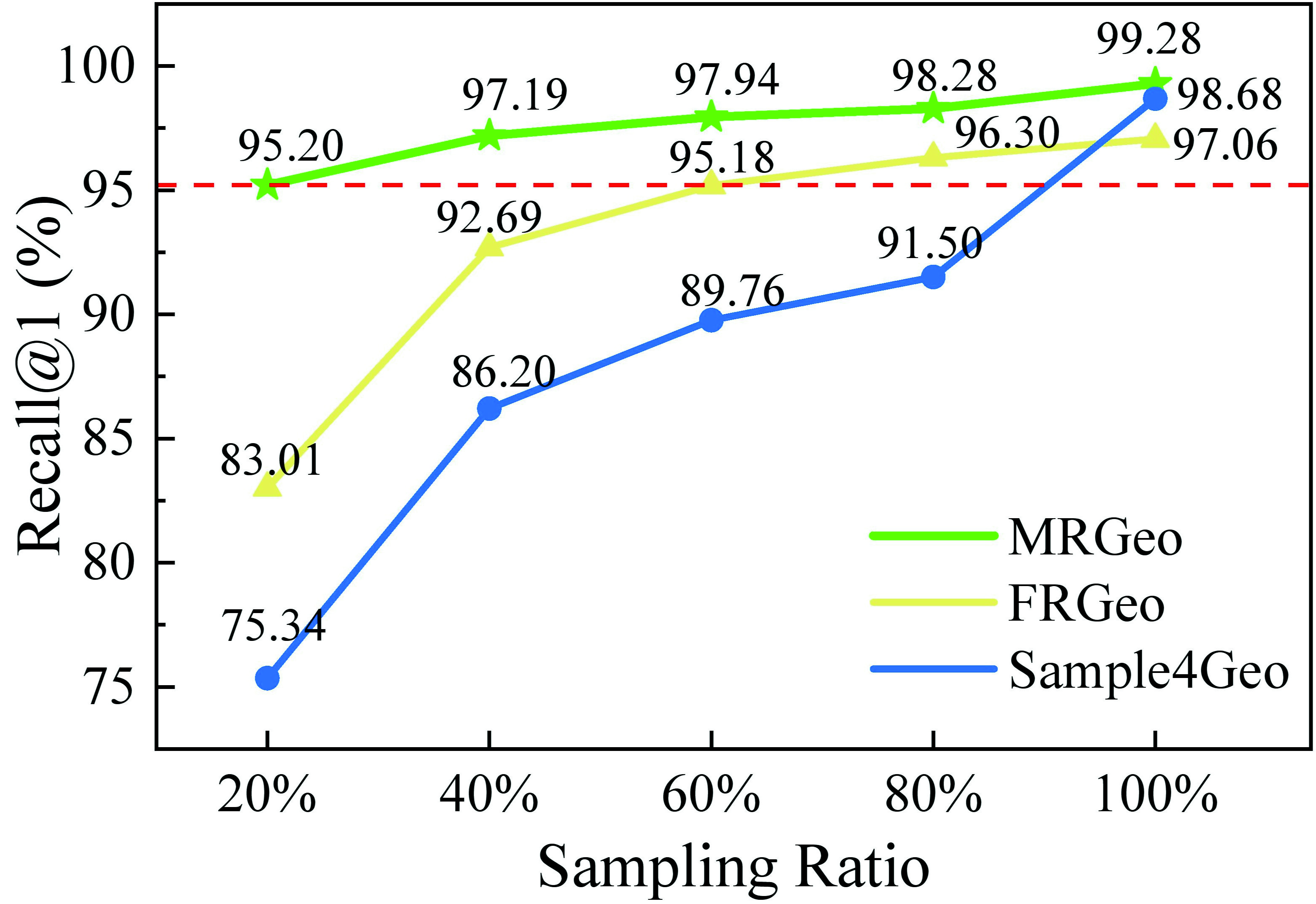}
                \caption{Few-shot training on CVUSA. Performance evaluation on the raw test set with progressively sampled training subsets (20\%, 40\%, 60\%, 80\%, 100\%). The red dotted line is MRGeo's R@1 benchmark on 20\% data.}
                \label{fig:few_shot_radar}
            \end{figure}
    
            \paragraph{Few-Shot Training.} We evaluate MRGeo's learning capability via few-shot learning to test its performance with limited data. As shown in Fig. \ref{fig:few_shot_radar}, MRGeo demonstrates exceptional efficiency. Using just 20\% of the training data, our model achieves an R@1 of 95.2\%. This result is not only significantly better than Sample4Geo using 80\% of the data (R@1 91.5\%) but also eclipses FRGeo's performance with 60\% of the data (R@1 95.18\%). This highlights MRGeo's ability to learn robust representations from limited information, making it a highly efficient and practical solution for scenarios where data acquisition is costly or challenging.
              
            \begin{figure}[t]
                \centering
                \includegraphics[width=1\linewidth]{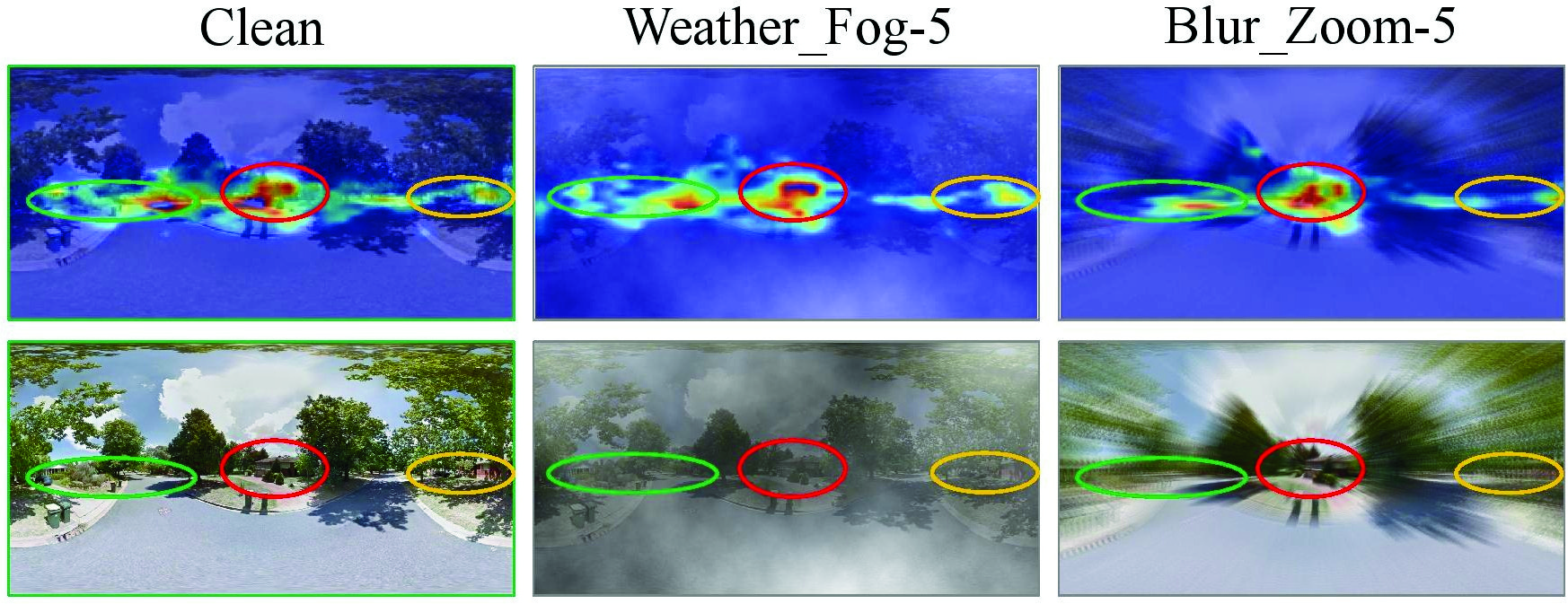}
                \caption{Heatmap visualization of MRGeo's feature focus on both clean and corrupted images. Best viewed on screen with zoom-in.}
                \label{fig:visual_heatmap}
            \end{figure}
        
        \subsection{Visualization of Qualitative Results}
        
            To better demonstrate the robustness of MRGeo to corrupted images, we visualize heatmaps on clean and corrupted images. Under ‘‘Fog'' and ‘‘Zoom'' corruption (Severity-5), MRGeo still exhibits excellent feature extraction and analysis capabilities. Compared to focus areas on clean images, it fails only on some minor features affected by interference.

    \section{Conclusion and Limitations}
    \label{conclusion}
    
        This paper proposes \textbf{MRGeo}, the first systematic method to address the performance degradation of cross-view geo-localization models on corrupted images. By building a hierarchical defense—dynamically enhancing features at the spatial and channel level with our \textbf{SARM} and \textbf{CCM}, and enforcing structural consistency at the descriptor level with \textbf{RGAM}—MRGeo not only sets a new SOTA on robustness benchmarks but also exhibits superior generalization ability in cross-area evaluation and high sample efficiency in few-shot training. These results validate our core thesis that true robustness stems from a holistic, multi-level enhancement strategy. However, our method's performance degrades under extremely severe corruptions, as unstructured noise can violate modules' assumptions and RGAM's effectiveness relies on a strict center-alignment. Addressing these challenges remains an important direction for future research.

    \section*{Acknowledgments}
        This work was supported in part by Shenzhen Science and Technology Program under Grant JCYJ20240813142510014 and Grant 20220810142553001,  in part by the Key Project of Department of Education of Guangdong Province under Grant 2023ZDZX1016.

\bibliography{aaai2026}

\end{document}